\documentclass[conference]{IEEEtran}
\usepackage{amsmath,graphicx}
\usepackage{subfigure}
\usepackage{multirow}
\usepackage{palatino}
\usepackage{tikz}
\usepackage{bm}
\usetikzlibrary{shapes.geometric, arrows}

\begin{document}
%
\title{Sub-GMN: The Neural Subgraph Matching Network Model}

\author{\IEEEauthorblockN{Zixun Lan\IEEEauthorrefmark{3} \;\;\; Limin Yu\IEEEauthorrefmark{1} \;\;\; Linglong Yuan\IEEEauthorrefmark{2} \;\;\;
Zili Wu\IEEEauthorrefmark{3} \;\;\; Qiang Niu\IEEEauthorrefmark{3} \;\;\; Fei Ma\IEEEauthorrefmark{3}}
\IEEEauthorblockA{\IEEEauthorrefmark{1}Department of Electrical and Electronic Engineering, Xi'an Jiaotong-Liverpool University, Suzhou, China.}
\IEEEauthorblockA{\IEEEauthorrefmark{2}Department of Mathematical Sciences, The University of Liverpool, United Kingdom.}
\IEEEauthorblockA{\IEEEauthorrefmark{3}Department of Applied Mathematics, School of Science, Xi’an Jiaotong-Liverpool University, Suzhou, 215123, China \\Telephone: (86) 512 8816 1633, Email: fei.ma@xjtlu.edu.cn}}

\maketitle

\begin{abstract}
As one of the most fundamental tasks in graph theory, subgraph matching is a crucial task in many fields, ranging from information retrieval, computer vision, biology, chemistry and natural language processing. Yet subgraph matching problem remains to be an NP-complete problem. This study proposes an end-to-end learning-based approximate method for subgraph matching task, called subgraph matching network (Sub-GMN). The proposed Sub-GMN firstly uses graph representation learning to map nodes to node-level embedding. It then combines metric learning and attention mechanisms to model the relationship between matched nodes in the data graph and query graph. To test the performance of the proposed method, we applied our method on two databases. We used two existing methods, GNN and FGNN as baseline for comparison. Our experiment shows that, on dataset 1, on average the accuracy of Sub-GMN are 12.21\% and 3.2\%  higher than that of GNN and FGNN respectively. On average running time Sub-GMN runs 20-40 times faster than FGNN. In addition, the average F1-score of Sub-GMN on all experiments with dataset 2 reached 0.95, which demonstrates that Sub-GMN outputs more correct node-to-node matches.

Comparing with the previous GNNs-based methods for subgraph matching task, our proposed Sub-GMN allows varying query and data graphes in the test/application stage, while most previous GNNs-based methods can only find a matched subgraph in the data graph during the test/application for the same query graph used in the training stage. Another advantage of our proposed Sub-GMN is that it can output a list of node-to-node matches, while most existing end-to-end GNNs based methods cannot provide the matched node pairs.

\end{abstract}

\IEEEpeerreviewmaketitle

\section{Introduction}
Given a data graph $G$ and a query graph $Q$, the subgraph matching problem \cite{liu2019g} is to find a mapping $m$ that is an isomorphism between $Q$ and a subgraph of $G$.

Subgraph matching has been widely used in various fields, ranging from information retrieval, computer vision, and natural language processing. In information retrieval, a query process of searching for research papers is a subgraph matching task \cite{hong2015subgraph}. In computer vision, through some preset rules, images and objects can be converted into graphs, and object recognition can be treated as a subgraph matching problem \cite{llados2001symbol}. In natural language processing, words in the corpus can be regarded as nodes, and the relationship between words can be viewed as edges, so paraphrases can be equivalent to subgraph matching \cite{nastase2015survey}.

Generally, subgraph matching is an NP-hard problem. Some exact subgraph matching methods have been proposed, such as Ullman’s \cite{ullmann1976algorithm}, vf2 \cite{cordella2004sub}, Ceci \cite{bhattarai2019ceci}, FilM \cite{moorman2018filtering}, etc., all of which have exponential time complexity in worst cases. The number of nodes is huge in real world graphs, so the time required for exact matching is huge. In addition to exponential time complexity, the graphs in the reality often have noise, which may cause the data graph not to contain the exact matching subgraph and results in returning nothing after a long time calculation. To solve the subgraph matching problem within a reasonable time and with a noisy background, one has to look for a fast inexact method tolerating existence of noise.

Existing inexact subgraph matching methods include Saga \cite{tian2007saga}, Tale \cite{tian2008tale}, G-Ray \cite{tong2007fast} and so on.
Most inexact methods rely on certain heuristics to select appropriate seed nodes, and then expand to neighbors of seeds with preset rules. For example, G-Finder \cite{liu2019g} chooses the node $u$ as the root node, where $u$ satisfies the conditions of the largest degree of node and smallest candidate set composed of nodes with the same feature as $u$ in data graph. After that, it continuously expands nodes from the root node according to preset rules to build LOOKUP-TABLE GRAPH. Finally, the enumeration strategy is executed in the set of candidate nodes until the k best subgraphs are selected. G-Ray \cite{tong2007fast} propose an intuitive goodness score $g$ to measure how well a subgraph matches the query pattern and a fast algorithm to find and rank qualified subgraphs. Saga \cite{tian2007saga} employs a flexible graph distance model to measure similarities between graphs and identify subgraphs. Contrary to the idea of heuristics and preset rules, the idea of deep learning is to obtain underlying knowledge and information based on learning from a large number of samples. In other words, the potential information of data graph and query graph can be used in subgraph matching tasks.

Deep learning has also been applied to non-Euclidean data gradually, such as graph, after it was maturely used in Euclidean data, such as pictures and words. For graph data, there are some graph representation learning methods, such as GNN \cite{scarselli2008graph}, GCN \cite{kipf2016semi}, graphSAGE \cite{hamilton2017inductive}, GAT \cite{velivckovic2017graph}, GIM \cite{xu2018powerful}, and so on. These methods essentially map high-dimensional structural data to low-dimensional embedding spaces, and represent graphs, edges and nodes with low-dimensional embedding that are often sent to a downstream learner for tasks of classification or regression. Downstream tasks usually have three categories, namely node classification \cite{kipf2016semi} in node-level, link prediction \cite{zhang2018link} in edge-level, graph classification \cite{rong2019dropedge} or graph similarity calculation \cite{li2019graph} in graph-level.

Although GNNs achieve good performance in many similarity measurement tasks \cite{bai2019simgnn, lan2022more}, such as measuring the similarity between two different graphs, they are not good at substructure extraction tasks, such as shortest path, subgraph, self-loop, etc. The methods using GNNs for tasks related to subgraph matching include GNN  \cite{scarselli2008graph}, FGNN \cite{krlevza2016graph}, Counting \cite{liu2020neural}. Among them, the main purpose of Counting is not to find out the query graph from the data graph, but to count the number of query graphs in the data graph. GNN \cite{scarselli2008graph} and FGNN \cite{krlevza2016graph} are essentially binary classification on nodes. They predict nodes in data graph that are also in query graph as +1, and nodes in data graph that are not in query graph as -1. The structure of their models has two unavoidable limitations in the subgraph matching task. On the one hand, the output of their models cannot get the node-to-node matching relationship, that is, only the position of the query graph on the data graph can be obtained, due to the binary classification on nodes. On the other hand, they can only match a fixed and unchanged subgraph, and they cannot match a different query graph in the test phase, because their input is a single data graph in the train and test phases. In order to break through the above two limitations, we need to change the input part of the model and enhance the expression of the node-to-node relationship between pairs of nodes.

In this paper, we propose a novel subgraph matching model called SubGraph Matching Network (Sub-GMN). Instead of a single data graph, the input of Sub-GMN is a pair of graphs consisting of a query graph and a data graph. This enables the model to match subgraph between different query graphs and different data graphes in the real application, rather than a fixed query graph. After obtaining the query graphs and data graph’s nodel-level embedding of each layer of GCN \cite{kipf2016semi}, by using neural tensor network \cite{socher2013reasoning}, node-to-node attention mechanism, Sub-GMN calculates the similarity among pairs of nodes. Finally, the model outputs the weighted average of the similarities obtained at each layer of GCN as a matching matrix that can represent the node-to-node matching relationship \cite{shen2022dynamical}.

We use the same data generation method as used in \cite{scarselli2008graph} and \cite{krlevza2016graph} to generate experimental data. From the experiment Sub-GMN demonstrated better performance compared to previous learning based GNNs models \cite{scarselli2008graph,krlevza2016graph}  in terms of accuracy and running time. The proposed Sub-GMN outputs node-to-node matching relationships. To our best knowledge, no learning based methods have been proposed in formal journals that match subgraphs, and output node-to-node matching relationships, while allow varying query and data graphes.

The rest of this paper is organized as follows. The preliminary is introduced in section \ref{sec_Preliminary}, then the model is described in section \ref{sec_Model}, and finally the experimental result is presented in section \ref{sec_Experiment}. Conclusion is made in section \ref{sec_Conclusion}.



\section{Preliminary}\label{sec_Preliminary}
\subsection{Subgraph matching and Matching Matrix}
A graph is denoted as a tuple \{$V$, $E$, $F$\}, where $V$ is the node set, $E$ is the edge set and $F$ is a feature function which maps a node or an edge to an feature vector. In this paper, $Q$ represents query graph, $G$ stands for data graph \cite{shen2020fabricated}.

Definition 1: Subgraph matching \cite{lan2022aednet}. The subgraph matching is an injective function $m: V(Q) \rightarrow V(G)$, which satisfies:
\begin{align*}
 & (1)\ \forall u \in V(Q), m(u) \in V(G)\;and\;F(u)=F(m(u));\\
 & (2)\ \forall\left(u_{a}, u_{b}\right) \in E(Q),\left(m\left(u_{a}\right), m\left(u_{b}\right)\right) \in E(G)\;and\;\\
 & F\left(u_{a}, u_{b}\right)=F\left(m\left(u_{a}\right), m\left(u_{b}\right)\right),\\
 \end{align*}
where $F$ is feature function.

Definition 2: Matching Matrix. Matching Matrix contains the node-to-node matching relationship. Each row of it represents a node in the query graph, and each column represents a node in the data graph. The Matching Matrix $M \in R^{n \times m}$ ($n$ and $m$ are the sizes of $Q$ and $G$) is denoted as follows:
\begin{equation}
M_{i j}=\left\{\begin{array}{ll}
1 & m(i)=j \\
0 & m(i) \neq j
\end{array}\right.
\end{equation}
where $i$, $j$ represent nodes in $Q$ and $G$, $m$ is an injective function for subgraph matching. Therefore, each element in it represents whether there is a match between two nodes. In the training phase, Matching Matrix is also the supervision information used by our end-to-end model, because we expect the output of our model to be close enough to the ground-truth Matching Matrix.

\subsection{Graph Convolutional Networks}
In Sub-GMN, computing node-to-node similarity and performing attention mechanism require node-level embedding \cite{scarselli2008graph} that is representation vector for each node. Among many existing node embedding algorithms, we choose to use Graph Convolutional Networks (GCN) \cite{kipf2016semi}, because not only GCN is graph representation-invariant and inductive for any unseen graphs in test set, but also GCN is the most concise spatial models for nodes embedding. From the perspective of effect, it is also feasible to replace GCN with other strong enough graph representation learning.

The GCN model uses the structure of the graph to aggregate the neighbor node information, and then updates the node representation through a non-linear activation function. Its core operation, the spatial graph convolution layer, is denoted as follows:
\begin{equation}
H^{l+1}=f_{1}\left(\hat{A} H^{l} W_{1}^{l}\right)
\end{equation}
where $\hat{A}=\widetilde{D}^{-\frac{1}{2}} \widetilde{A} \widetilde{D}^{-\frac{1}{2}}\left(\widetilde{A}=A+I_{N}, \widetilde{D}_{i i}=\sum \widetilde{A}_{i i}\right) \in R^{N \times N}$ ($N$ is the number of nodes) is a normalized adjacency matrix, $H^{l} \in R^{N \times D^{\prime}}$ is node-level embedding of $l$-th GCN layer ($H^0=X$, $X$ is nodes feature matrix ), $W_{1}^{l} \in R^{D^{l} \times D^{l+1}}$ is a learnable weights of $l$-th GCN layer, and $f_{1}$ is a non-linear activation function.

\subsection{Neural Tensor Network}
The Neural Tensor Network (NTN) \cite{socher2013reasoning} is suitable for reasoning over relationships between two entities. In Sub-GMN, a pair of nodes including one in the query graph and the other one in the data graph can be calculated the similarity through NTN. The Neural Tensor Network (NTN) uses a bilinear tensor layer that directly relates the two entity vectors across multiple dimensions. The model computes a score of how likely it is that two entities are in a certain relationship by the following NTN-based function:
\begin{equation}
g\left(e_{1}, e_{2}\right)=f_{2}\left(e_{1}^{T} W_{2}^{[1: k]} e_{2}+V_{2}\left[\begin{array}{c}
e_{1} \\
e_{2}
\end{array}\right]+b_{2}\right)
\end{equation}
where $f_2$ is a non-linear activation function, $W_{2}^{[1: k]} \in R^{k \times d \times d}$, $V_{2} \in R^{k \times 2 d}$, $b_{2} \in R^{k}$ are learnable parameters, $e_{1}, e_{2} \in R^{d \times 1}$ are representations of two entities, $[\cdot]$ is concatenation operation. Therefore, k similarity scores are obtained through NTN between two entities.

\begin{figure*}[t]
\centering
\includegraphics[width=0.7\linewidth]{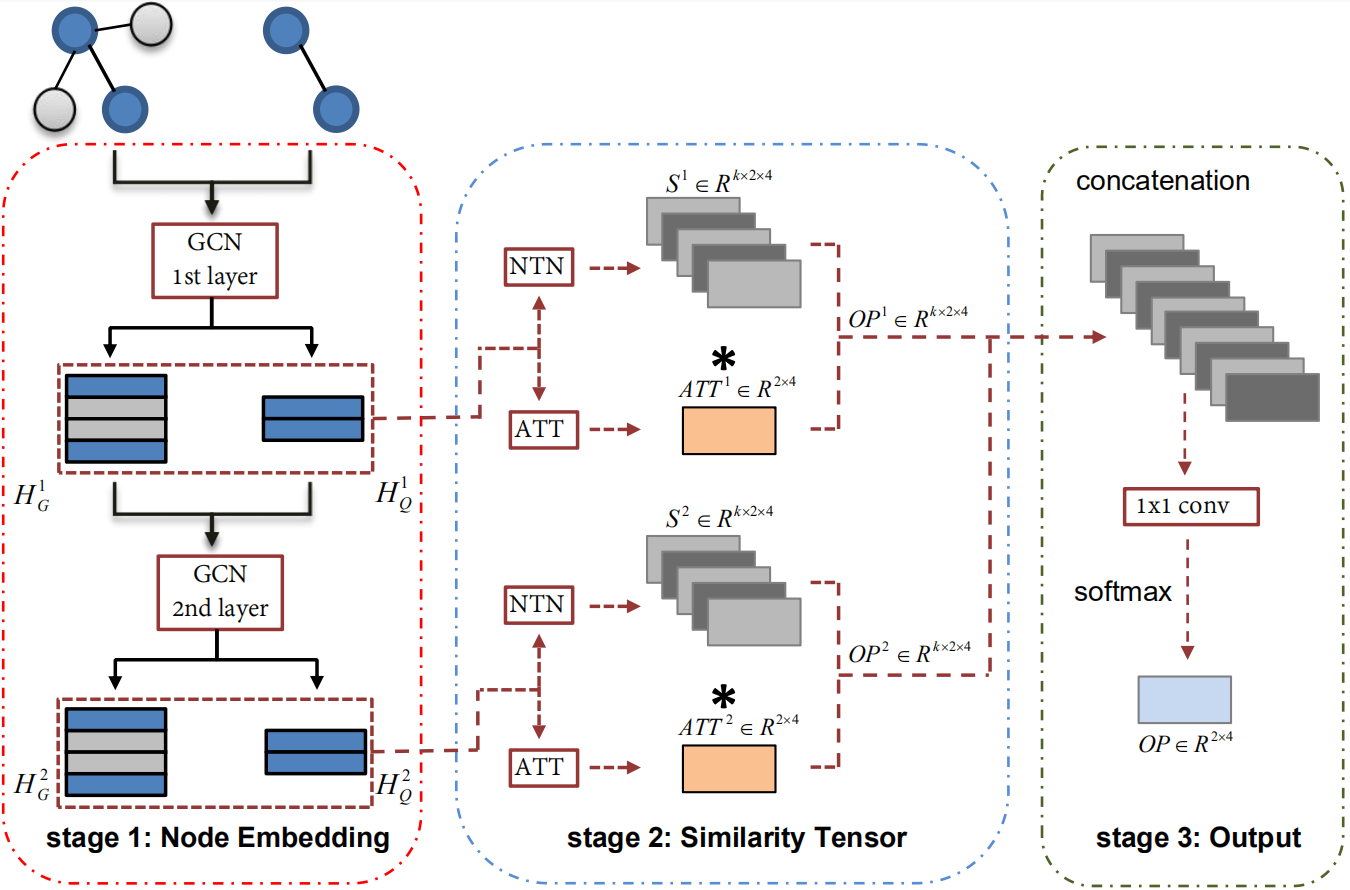}
\caption{An example of Sub-GMN of two layers. The red, blue and green dotted boxes represent stage 1, stage 2, and stage 3 respectively.}

\end{figure*}


\section{Model}\label{sec_Model}
The Sub-GMN is an end-to-end neural network that attempts to learn a function to map a pair of query graph and data graph into a predicted Matching Matrix based on the assumption that the higher similarity between pairs of nodes, the higher the probability of matching between pairs of nodes. An example’s overview of Sub-GMN is illustrated in Fig. 1. First, our model transforms nodes of each graph into a vector, encoding the features and structural information around each node by GCN in stage 1. In stage 2, in order to calculate the similarity that is used to construct predicted Matching Matrix between pairs of nodes, we use NTN and node-to-node attention mechanism together to get the similarity between pairs of nodes at the end of each layer of GCN. Finally, the model uses 1x1 convolution to reduce the dimension of the concatenated similarity tensors to the output of the model whose shape is the same as the shape of the ground-truth Match matrix in stage 3. The rest of the section details these three stages.

\subsection{Stage 1: Node Embedding}
Among the existing state-of-the-art approaches, we adopt GCN described in section 2, because not only GCN \cite{kipf2016semi} is graph representation-invariant and inductive for any unseen graphs in test set, but also GCN is the most concise spatial models for nodes embedding. The model obtains node-level embedding through the following formula:
\begin{equation}
H_{Q}^{l}=f_{1}\left(\hat{A}_{Q} H_{Q}^{l-1} W_{1}^{l-1}\right)
\end{equation}
\begin{equation}
H_{G}^{l}=f_{1}\left(\hat{A}_{G} H_{G}^{l-1} W_{1}^{l-1}\right)
\end{equation}
where, $H_{Q}^{l}$ and $H_{G}^{l}$ are the node-level embedding $Q$ and $G$ respectively, $f_1$ is a non-linear activation function. It is worth noting that the weight of GCN here is shared for $Q$ and $G$, because we expect the more similar nodes to have more similar embedding.

\subsection{Stage 2: Similarity tensor}
Given the embedding of two nodes produced by the previous stage, a simple way to compute their similarity is to take the inner product between them. However, such simple usage of data representations often creates weak connection between pairs of nodes. Therefore, we use Neural Tensor Networks (NTN) \cite{socher2013reasoning} combined with node-to-node attention mechanism to model the relationship between two nodes:
\begin{equation}
S_{i j}^{[1: k]^{l}}=f_{2}
\left(H_{Q_{i}}^{l} W_{2}^{[1: k]} {H_{G_{j}}^{l}}^{T}+V_{2}\left[\begin{array}{c}
{H_{Q_{i}}^{l}}^{T} \\
{H_{G_{j}}^{l}}^{T}
\end{array}\right]+b_{2}\right)
\end{equation}

\begin{equation}
A T T^{l}=\sigma\left(f_{2}\left(\frac{H_{Q}^{l} {H_{G}^{l}}^{T}}{\sqrt{D^{l}}}\right)\right)
\end{equation}
\begin{equation}
O P^{[1: k]^{l}}=S^{[1: k]^{l}} * A T T^{l}
\end{equation}
where $H_{Q_{i}}^{l}$, $H_{G_{j}}^{l} \in R^{1 \times D}$ are embedding of one node in $Q$ and $G$, $A T T^{l} \in R^{n \times m}$ is node-to-node attention ($n$ and $m$ are the sizes of $Q$ and $G$),  $S^{l} \in R^{k \times n \times m}$ is node to node similarity after each layer of GCN, $f_{2}$ is a sigmoid function, $\sigma$ is a softmax function which ensure that the sum of each row of $ATT^{l}$ is one and $O P^{l} \in R^{k \times n \times m}$ is the similarity tensor at the end of l-th layer of GCN.

\subsection{Stage 3: Output and Loss function}
After stage 1 and stage 2, the model obtain l $OP^{l}$s and perform concatenation operation on these. The model uses $1 \times 1$ convolution to reduce the dimension of the concatenated similarity tensors to the output of the model whose shape is the same as the shape of the ground-truth Match matrix:
\begin{equation}
O P=\sigma\left(\operatorname{Conv}_{1 \times 1}\left(\operatorname{concatenation}\left(O P^{1}, \cdots, O P^{l}\right)\right)\right)
\end{equation}
where $O P \in R^{n \times m}$ is the final output of Sub-GMN, $\sigma$ is a softmax function which ensure that the sum of each row of $OP$ is one. During the train process, it is compared against the ground-truth Matching Matrix $M$ using the following loss function:
\begin{equation}
L=\frac{1}{|D|} \sum\|O P-M\|_{F}
\end{equation}
where $D$ is the training set and $|D|$ represents the number of samples in training set $|D|$. In the test phase, we set the largest element of each row of $OP$ to 1, and the rest to 0, because the output of the model is continuous, while the elements of the matching matrix are discrete.

\section{Experiment}\label{sec_Experiment}
In this section, from the perspective of effectiveness and efficiency, we carry out experiments by comparing subgraph matching GNNs-based methods.

\subsection{Datasets}
Our experiment is carried out on two datasets. Dataset 1 is generated using the same method as used in GNN \cite{scarselli2008graph} and FGNN \cite{krlevza2016graph}, which means that the query graph in each sample pair of graphes is identical and unchanged in dataset 1. On the contrary, the data graph and the query graph of each sample graph pair are different in dataset 2. In the experiment, we use an undirected graph generator as used in \cite{scarselli2008graph}. The generator has three parameters, including the size $|g|$ of the graph to be generated, the probability $p$ of generating an edge between node and node, and the maximum value for the node feature $N$.

Each node is assigned a random integer in range $[1, N]$ as the node feature. In this study, the parameter $p$ of undirected graph generator is set to 0.2, which is the same as that in \cite{scarselli2008graph} and \cite{krlevza2016graph}. In order to test the ability of the model in the presence of noise, we add a Gaussian noise with mean value of 0 and variance of 0.25 to each node feature. The details of dataset are presented in table \ref{t1}.

\begin{table*}[t]
\begin{center}
\caption{Details of dataset. $|G|$ and $|Q|$ are size 0f data graph and query graph respectively, $|N|$ is the maximum value for the node feature.}
\setlength{\tabcolsep}{5.0mm}
\label{t1}
\begin{tabular}{c c c c c c }\hline\hline
 $Dataset$  & query graph &$|G|$&$|Q|$&$|N|$&size of train$\backslash$validation$\backslash$test set\\  \hline
  Dataset1	&	unchanged     &	6, 10, 14, 18	    &	3, 5, 7, 9	 &10& 200$\backslash$200$\backslash$200 \\
Dataset2	&changed	&6, 9, 18, 300&3, 6, 9, 10&10, 40& 8000$\backslash$1000$\backslash$200 \\ \hline\hline
\end{tabular}
\end{center}
\end{table*}

\begin{table*}[t]
\begin{center}
\caption{Results on dataset 1 between Sub-GMN and GNN. $|G|$ and $|Q|$ are size 0f data graph and query graph respectively, $|N|$ is the maximum value for the node feature.}
\setlength{\tabcolsep}{1.0mm}
\label{t2}
\begin{tabular}{c c c c c c c c c c c c c c c}\hline\hline
 $(|G|, |Q|, |N|=10)$  & (18, 3) & (18, 5) &	(18, 7) &(18, 9)&(14, 3)&(14, 5)&(14, 7)&(14, 9)&(10, 3)&(10, 5)&(10, 7)&(10, 9)&(6, 3) &(6, 5)\\  \hline
  Accuracy of GNN	&	84.30\%	        &	83.30\%	    &	79.90\%	 &77.80\%&90.00\%&84.90\%&84.60\%&84.00\%&90.00\%&87.70\%&89.80\%&93.30\%&92.40\%&91.30 \% \\
Accuracy of Sub-GMN	&	\textbf{99.23\%}  &	\textbf{98.92\%} &	\textbf{98.56\%}	 &\textbf{98.23\%}&\textbf{99.17\%}&\textbf{98.56\%}&\textbf{98.65\%}&\textbf{98.84\%}&\textbf{98.53\%}&\textbf{96.84\%}&\textbf{99.35\%}&\textbf{99.73\%}&\textbf{99.93\%}&\textbf{99.66\%}  \\ \hline\hline
\end{tabular}
\end{center}
\end{table*}

\begin{table*}[t]
\begin{center}
\caption{Results on dataset 2. $|G|$ and $|Q|$ are size 0f data graph and query graph respectively, $|N|$ is the maximum value for the node feature.}
\setlength{\tabcolsep}{1.0mm}
\label{t3}
\begin{tabular}{c c c c c c c c c c c c c c c}\hline\hline
 $(|G|, |Q|, |N|=10)$  & (18, 3) & (18, 5) &	(18, 7) &(18, 9)&(14, 3)&(14, 5)&(14, 7)&(14, 9)&(10, 3)&(10, 5)&(10, 7)&(10, 9)&(6, 3) &(6, 5)\\  \hline
  Accuracy of Sub-GMN	&	97.12\%	        &	94.65\%	    &	94.83\%	 &94.28\%&98.29\%&95.43\%&94.93\%&97.07\%&99.20\%&96.10\%&97.60\%&99.10\%&98.83\%&99.17\% \\
F1 for Sub-GMN	&0.93	&0.87&0.93&0.92&0.95&0.93&0.92&0.96&0.98&0.94&0.97&0.99&0.99&0.99 \\ \hline\hline
\end{tabular}
\end{center}
\end{table*}

\subsubsection{Dataset 1}
We first generate a query graph $Q$ whose size is $|Q|$ through the generator, and then generate $D_1$ graphs $g_i$ whose sizes are $|G|-|Q|$, where $G$ is the data graph. Finally, we randomly insert this query graph $Q$ into each $g_i$, so that we obtain $D_1$ samples and each sample is a graph pair containing a data graph and a query graph. In dataset 1, the query graph of each sample is identical, except some small difference in node feature duo to noise. In our first experiment, $D_1$ is 600, where there are 200, 200, 200 examples in the training set, validation set and test set respectively.

\subsubsection{Dataset 2}
We generate $D_2$ query graphs $Q_i$ ($i=1,2, \cdots ,D_2$) whose sizes are $|Q|$ through the generator, and then generate $D_2$ graphs $g_i$ of size $|G|-|Q|$, where $G$ is the data graph. Finally, we randomly insert $D_2$ query graphs $Q_i$ into each $g_i$ in turn, so that we get $D_2$ samples and each sample is a graph pair. In dataset 2, the query graph in each sample is different. In our second experiment, $D_2$ is 10000, where there are 8000, 1000, 1000 examples in the training set, validation set and test set respectively.

\subsection{Baseline Methods}
Two subgraph matching GNNs-based methods are used as the baselines, including GNN \cite{scarselli2008graph} and FGNN \cite{krlevza2016graph}, because these two models are specifically for subgraph matching tasks. Although there are other GNNs-based methods for graph matching and graph similarity calculations, such as GMN \cite{li2019graph}, Simgnn \cite{bai2019simgnn}, they are all for graph matching tasks, not for subgraph matching tasks.

These two baseline methods can only predict which nodes in the data graph are nodes in the query graph, and cannot output node to node matching relationship. On the other hand, they can only predict a fixed query graph appearing in the train set. In contrast, Sub-GMN can predict the node-to-node matching relationship, and it can also match different query graphs during the test phase.

\subsection{Parameters Setting}
For the model architecture, we set the number of GCN layers to 3, use elu as the first and second layers’ activation functions and use softmax as the third layer’s activation function. The output dimensions for the 1st, 2nd, and 3rd layer of GCN are 128, 128, and 128, respectively. For the NTN layer, we set K to 16.

We conduct all the experiments on a single machine with an Intel Xeon 4114 CPU and one Nvidia Titan GPU. As for training, we set the batch size to 128, use the Adam algorithm for optimization \cite{kingma2014adam} and fix the initial learning rate to 0.001. We set the number of iterations to 5000, and select the best model based on the lowest validation loss.

\subsection{Evaluation Metrics}
We use the accuracy, F1-Score and running time to evaluate the node classification performace, the node to node matching accuracy and the efficiency.

\subsubsection{Accuracy}
Our experiment used the same accuracy criteria as used in the experiment in \cite{scarselli2008graph}, that is, the ratio of the number of nodes correctly classified in each data graph to the total number of nodes in data graph:
\begin{equation}
\text { accuracy }=\frac{NOCC}{TNON}
\end{equation}
where $NOCC$ and $TNON$ represent the number of nodes correctly classified and the total number of nodes in data graph respectively. Under this evaluation metric, all nodes in data graph do a binary classification to identify whether this node is a node in the query graph.

\subsubsection{F1-Score}
\begin{equation}
F_{1}=\frac{2 \cdot \text {P} \cdot \text {R}}{\text {P}+\text {R}}
\end{equation}
where $P$ is precision representing the ratio of the number of correctly discovered node matches over number of all discovered node matches, $R$ is recall representing the ratio of the correctly discovered node matches over all correct node matches.

\subsubsection{Running Time}
We also use the running time to evaluate the efficiency of models.

\subsection{Results}
Table \ref{t2} and \ref{t4} list 14 and 4 settings of sample graph pair respectively. These settings are selected as they are also used in \cite{scarselli2008graph} and \cite{krlevza2016graph}. This paper used these same setting for the convenience of comparison.
\subsubsection{Compared with GNN}
It can be seen from table \ref{t2} that the accuracy of all experimental results of Sub-GMN in dataset1 outperform GNN, and the accuracy of Sub-GMN is 12.21\% higher than that of GNN on average. It is worth noting that the accuracy of GNN varies greatly in different tasks of dataset 1, while the accuracy of Sub-GMN is basically concentrated around 98\% on dataset1, which means that Sub-GMN has the ability to find the position of the query graph in the data graph and is more powerful and stable than GNN for subgraph matching problem.

\subsubsection{Compared with FGNN}
From table \ref{t4}, although FGNN performs well with an average accuracy of 96.03\% on dataset 2, the accuracy of Sub-GMN is still 3.2\% higher than that of FGNN on average. FGNN's accuracy on the third task of dataset2 is only 89.9\%. The possible reason is that the size of data graph and the size of query graph are quite different, so that the matching relationship cannot be identified by node features. On the contrary, the accuracy of Sub-GMN is basically concentrated around 99\% on dataset2, which means that Sub-GMN is more powerful and stable than FGNN for subgraph matching problem and can study more information for structure of graphs. On the other hand, It can be seen from table \ref{t5} that the average running time of Sub-GMN runs 20-40 times faster than FGNN.

\subsubsection{Results on dataset 2}
From table \ref{t3}, the average accuracy of all experimental results in dataset 2 reached 96.9\%, which shows that it can match different query graphs during the test phase. On the other hand, the average F1-score of all experimental results in dataset2 reached 0.95, which demonstrates that Sub-GMN can predict more exact node-to-node matching relationship.

\begin{table}[h]
\begin{center}
\caption{Results on dataset 1 between Sub-GMN and FGNN. $|G|$ and $|Q|$ are size 0f data graph and query graph respectively, $|N|$ is the maximum value for the node feature.}
\setlength{\tabcolsep}{1.0mm}
\label{t4}
\begin{tabular}{c c c c c }\hline\hline
 $(|G|, |Q|, |N|)$  & (6, 3, 10) & (9, 6, 10) &	(18, 9, 10) &(300, 10, 40)\\  \hline
  Accuracy of FGNN&	98.10\%	        &	97.90\%	    &	89.90\%	 &98.20\% \\
Accuracy of Sub-GMN	&\textbf{99.93\%}	&\textbf{99.21\%}&\textbf{98.23\%}&\textbf{99.56\%} \\ \hline\hline
\end{tabular}
\end{center}
\end{table}

\begin{table}[h]
\begin{center}
\caption{Average running time between Sub-GMN and FGNN. $|G|$ and $|Q|$ are size 0f data graph and query graph respectively, $|N|$ is the maximum value for the node feature.}
\setlength{\tabcolsep}{1.5mm}
\label{t5}
\begin{tabular}{c c c c c }\hline\hline
 $(|G|, |Q|, |N|)$  & (6, 3, 10) & (9, 6, 10) &	(18, 9, 10) &(300, 10, 40)\\  \hline
FGNN&	6ms        &	60ms	    &	80ms	 &300ms \\
Sub-GMN	&1ms	&1ms&2ms&35ms \\ \hline\hline
\end{tabular}
\end{center}
\end{table}

\section{Conclusion}\label{sec_Conclusion}
Sub-GMN is an end-to-end neural network that attempts to learn a function to map a pair of query graph and data graph into a predicted Matching Matrix. By using neural tensor network \cite{socher2013reasoning}, node-to-node attention mechanism, Sub-GMN can output the predicted matching matrix containing node-to-node matching relationships.

The experimental results demonstrate that our model outperforms other subgraph matching GNNs-based methods in performance and running time. It is worth mentioning that compared with the previous subgraph matching GNNs-based method, we can get the node-to-node matching relationship, and the F1 score of this is very high. However, there are two directions to go for the future work: (1) the scale of graph data in practical applications is huge, and this is also the next goal that this model is to be applied to large-scale graph with improved accuracy; (2) introducing the cross-graph propagation mechanism \cite{li2019graph,wang2019learning}, and representing the input pair of graphs as a relational whole.

\section{Acknowledgements}
This study was supported in part by XJTLU laboratory for intelligent computation and financial technology through XJTLU Key Programme Special Fund (KSF-P-02 and KSF-E-21).

\bibliographystyle{IEEEtran}
\bibliography{ree}

\end{document}